\documentclass[conference]{IEEEtran}
\makeatletter\if@twocolumn\PassOptionsToPackage{switch}{lineno}\else\fi\makeatother

      \makeatletter
\usepackage{wrapfig}
\newcounter{aubio}

\long\def\bioItem{%
\@ifnextchar[{\@bioItem}{\@@bioItem}}

\long\def\@bioItem[#1]#2#3{
 \stepcounter{aubio}
 \expandafter\gdef\csname authorImage\theaubio\endcsname{#1}
 \expandafter\gdef\csname authorName\theaubio\endcsname{#2}
 \expandafter\gdef\csname authorDetails\theaubio\endcsname{#3}
}

\long\def\@@bioItem#1#2{
 \stepcounter{aubio}
 \expandafter\gdef\csname authorName\theaubio\endcsname{#1}
 \expandafter\gdef\csname authorDetails\theaubio\endcsname{#2}
}

\newcommand{\checkheight}[1]{%
  \par \penalty-100\begingroup%
  \setbox8=\hbox{#1}%
  \setlength{\dimen@}{\ht8}%
  \dimen@ii\pagegoal \advance\dimen@ii-\pagetotal
  \ifdim \dimen@>\dimen@ii
    \break
  \fi\endgroup}

\def\printBio{%
  \@tempcnta=0
   \loop
     \advance \@tempcnta by 1
     \def\aubioCnt{\the\@tempcnta}
     \setlength{\intextsep}{0pt}%
     \setlength{\columnsep}{10pt}%
     \newbox\boxa%
     \setbox\boxa\vbox{\csname authorDetails\aubioCnt\endcsname}
     \expandafter\ifx\csname authorImage\aubioCnt\endcsname\relax%
      \else%
       \checkheight{\includegraphics[height=1.25in,width=1in,keepaspectratio]{\csname authorImage\aubioCnt\endcsname}}
        \begin{wrapfigure}{l}{25mm}
         \includegraphics[height=1.25in,width=1in,keepaspectratio]{\csname authorImage\aubioCnt\endcsname}
        \end{wrapfigure}\par
      \fi
     {\parindent0pt\textbf{\csname authorName\aubioCnt\endcsname}\csname authorDetails\aubioCnt\endcsname \par\bigskip%
     \expandafter\ifx\csname authorImage\aubioCnt\endcsname\relax\else%
      \ifdim\the\ht\boxa < 90pt\vskip\dimexpr(90pt -\the\ht\boxa-1pc)\fi%
     \fi}
      \ifnum\@tempcnta < \theaubio
   \repeat
   }

\makeatother

\usepackage{graphicx}
\usepackage[T1]{fontenc}
\ifCLASSINFOpdf
\else
\fi
\usepackage{amsmath}
\usepackage[cmintegrals]{newtxmath}
\usepackage{url,multirow,morefloats,floatflt,cancel,tfrupee}
\makeatletter

\AtBeginDocument{\@ifpackageloaded{textcomp}{}{\usepackage{textcomp}}}
\makeatother
\usepackage{colortbl}
\usepackage{xcolor}
\usepackage{pifont}
\usepackage[nointegrals]{wasysym}
\makeatletter

\def\mcWidth#1{\csname TY@F#1\endcsname+\tabcolsep}

\def\cAlignHack{\rightskip\@flushglue\leftskip\@flushglue\parindent\z@\parfillskip\z@skip}
\def\rAlignHack{\rightskip\z@skip\leftskip\@flushglue \parindent\z@\parfillskip\z@skip}

\@ifundefined{etal}{}{}

\usepackage{ifxetex}
\ifxetex\else\if@twocolumn\@ifpackageloaded{stfloats}{}{\usepackage{dblfloatfix}}\fi\fi

\AtBeginDocument{
\expandafter\ifx\csname eqalign\endcsname\relax
\def\eqalign#1{\null\vcenter{\def\\{\cr}\openup\jot\m@th
  \ialign{\strut$\displaystyle{##}$\hfil&$\displaystyle{{}##}$\hfil
      \crcr#1\crcr}}\,}
\fi
}

\AtBeginDocument{%
  \@ifpackageloaded{endfloat}%
   {\renewcommand\efloat@iwrite[1]{\immediate\expandafter\protected@write\csname efloat@post#1\endcsname{}}}{\newif\ifefloat@tables}%
}%

\def\BreakURLText#1{\@tfor\brk@tempa:=#1\do{\brk@tempa\hskip0pt}}
\let\lt=<
\let\gt=>
\def\processVert{\ifmmode|\else\textbar\fi}

\@ifundefined{subparagraph}{
\def\subparagraph{\@startsection{paragraph}{5}{2\parindent}{0ex plus 0.1ex minus 0.1ex}%
{0ex}{\normalfont\small\itshape}}%
}{}

\newcommand\role[1]{\unskip}
\newcommand\aucollab[1]{\unskip}
  
\@ifundefined{tsGraphicsScaleX}{\gdef\tsGraphicsScaleX{1}}{}
\@ifundefined{tsGraphicsScaleY}{\gdef\tsGraphicsScaleY{.9}}{}
\def\checkGraphicsWidth{\ifdim\Gin@nat@width>\linewidth
	\tsGraphicsScaleX\linewidth\else\Gin@nat@width\fi}

\def\checkGraphicsHeight{\ifdim\Gin@nat@height>.9\textheight
	\tsGraphicsScaleY\textheight\else\Gin@nat@height\fi}

\def\fixFloatSize#1{}
\let\ts@includegraphics\includegraphics

\def\inlinegraphic[#1]#2{{\edef\@tempa{#1}\edef\baseline@shift{\ifx\@tempa\@empty0\else#1\fi}\edef\tempZ{\the\numexpr(\numexpr(\baseline@shift*\f@size/100))}\protect\raisebox{\tempZ pt}{\ts@includegraphics{#2}}}}

\AtBeginDocument{\def\includegraphics{\@ifnextchar[{\ts@includegraphics}{\ts@includegraphics[width=\checkGraphicsWidth,height=\checkGraphicsHeight,keepaspectratio]}}}

\DeclareMathAlphabet{\mathpzc}{OT1}{pzc}{m}{it}

\def\URL#1#2{\@ifundefined{href}{#2}{\href{#1}{#2}}}

\def\UrlOrds{\do\*\do\-\do\~\do\'\do\"\do\-}%
\g@addto@macro{\UrlBreaks}{\UrlOrds}

\edef\fntEncoding{\f@encoding}

\makeatother

\newif\ifmultipleabstract\multipleabstractfalse%
\usepackage{tabulary}
\makeatletter
\AtBeginDocument{\@ifpackageloaded{longtable}{%
\def\LT@makecaption#1#2#3{%
  \LT@mcol\LT@cols c{\hbox to\z@{\hss\parbox[t]\LTcapwidth{%
    \sbox\@tempboxa{#1{#2: } #3}%
    \ifdim\wd\@tempboxa>\hsize
      #1{#2: }\textsc{#3}%
    \else
      \hbox to\hsize{\hfil\box\@tempboxa\hfil}%
    \fi
    \endgraf\vskip\baselineskip}%
  \hss}}}
}{}}
\makeatother

   \makeatletter
  \def\fig@textbf{\textbf}
   \AtBeginDocument{\renewcommand\floatc@plain[2]{\setbox\@tempboxa\hbox{{\footnotesize#1.}\footnotesize\hskip.5em#2}%
    \ifdim\wd\@tempboxa>\hsize {\fig@textbf{\footnotesize#1.}}\footnotesize\hskip.5em#2\par
        \else\hbox to\hsize{\hfil\box\@tempboxa\hfil}\fi}}
    \makeatother
  
\usepackage{float}

\begin{document}

%


        \title{AlphaNet: An Attention Guided Deep Network for Automatic Image Matting}
      \author{
		\IEEEauthorblockN{Rishab~Sharma}

    \IEEEauthorblockA{\textit{Fynd}}\\[-12pt]Email: rishabsharma@fynd.com ~\\(Corresponding author)
        \vspace*{1pc}\and 
		\IEEEauthorblockN{Rahul~Deora}

    \IEEEauthorblockA{\textit{Fynd}}\\[-12pt]Email: rahuldeora@fynd.com
        \vspace*{1pc}\and 
		\IEEEauthorblockN{Anirudha~Vishvakarma}

    \IEEEauthorblockA{\textit{Fynd}}\\[-12pt]Email: anirudhav@fynd.com}
  


\maketitle 

\begin{abstract}
In this paper, we propose an end to end solution for image matting i.e high-precision extraction of foreground objects from natural images. Image matting and background detection can be achieved easily through chroma keying in a studio setting when the background is either pure green or blue. Nonetheless, image matting in natural scenes with complex and uneven depth backgrounds remains a tedious task that requires human intervention. To achieve complete automatic foreground extraction in natural scenes, we propose a method that assimilates semantic segmentation and deep image matting processes into a single network to generate detailed semantic mattes for image composition task. The contribution of our proposed method is two-fold, firstly it can be interpreted as a fully automated semantic image matting method and secondly as a refinement of existing semantic segmentation models.

We propose a novel model architecture as a combination of segmentation and matting that unifies the function of upsampling and downsampling operators with the notion of attention. As shown in our work, attention guided downsampling and upsampling can extract high-quality boundary details, unlike other normal downsampling and upsampling techniques. For achieving the same, we utilized an attention guided encoder-decoder framework which does unsupervised learning for generating an attention map adaptively from the data to serve and direct the upsampling and downsampling operators. We also construct a fashion e-commerce focused dataset with high-quality alpha mattes to facilitate the training and evaluation for image matting.
\end{abstract}
    


%
\IEEEpeerreviewmaketitle

\section{Introduction}
Digital image matting, which is defined as high-quality extraction of foreground objects from natural images has a wide assortment of applications in mixed reality, film production, and smart creative composition. In an e-commerce website, image composition is usually used to generate smart and personalized creative assets for customers. Generating such compositions requires the extraction of fashion models and accessories from a large amount of fashion data, followed by their composition with the new creative backgrounds. In pursuit of smoother and faster customer experience, the process of extraction and composition that works on a huge volume of data must be automated and the results must be of high quality. An example of a creative composition in a real-world fashion e-commerce website achieved through automatic image matting can be seen in Fig. 1. For designing such an automated pipeline, one would either choose to go for image matting or semantic segmentation techniques. But this particular task is not as trivial as it seems, and there are many underlying bottlenecks that make it more complicated that one may perceive. Neither semantic segmentation nor image matting itself delivers a satisfactory result. One major drawback of using just semantic segmentation in this use case is that this technique focuses on coarse semantics of the visual input, thus leads to the blurring of fine structural details. On the contrary, simple image matting that is well-known for extracting fine details from images requires user interactions like trimaps or scribbles. Therefore, in such a time-sensitive and data-intensive scenario, only applying image matting cannot serve as a robust solution. In this work, we present a method for automatic foreground object extraction that can work even with complex natural backgrounds. Our method combines semantic segmentation and image matting processes, which allows for multiple foreground objects to be segmented and extracted from the background with high quality. The ability of the semantic segmentation to extract different types of semantic labels enables a way to automatically extract objects of different types.

\bgroup
\fixFloatSize{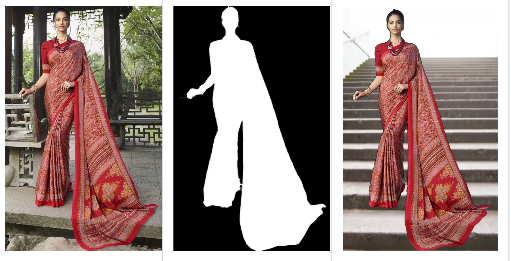}
\begin{figure}[t!]
\centering \makeatletter\IfFileExists{4d60d296-d226-4b99-be10-1ec3ff4deca5-ucollage.png}{\includegraphics{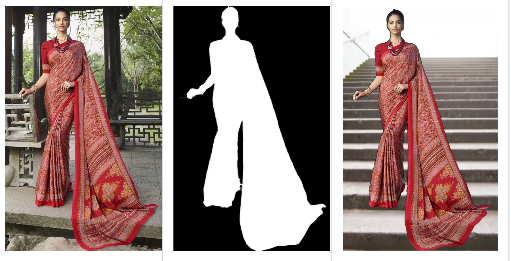}}{}
\makeatother 
\caption{{AlphaNet being used for an Fashion e-commerce website.}}
\label{f-7b3e729620b5}
\end{figure}
\egroup
One of the most complex and important tasks in computer vision is object segmentation. Although there have been recent advances in the learning-based methods for segmentation algorithms due to the availability of publicly available datasets, still most of the human drawn segmentation ground truths are unrefined and coarse. This coarse supervision makes the segmentation models generate unrefined and coarse object edges. Therefore, image segmentation techniques not only lack sufficient refinement but also many semantic details required for high-quality image composition. On the contrary, image matting is also an elementary problem in the domain of computer vision that is under continuous research since the 1950s. Identical to image segmentation, which generates a binary mask for the foreground, image matting extracts the foreground by estimating an alpha matte containing the transparency value for each pixel in the static image or a frame stream from a video. The pixel in the foreground typically has an alpha value of 1, whereas the background pixel has an alpha value of 0. Nonetheless, a pixel on the edge of an object can have a contribution from both the foreground and background due to motion blur in videos, thus the alpha is defined as $\alpha\in\lbrack0,1\rbrack $ in such a case. The interpreted relation of alpha and image can be seen in Equation 1, where $\alpha_i $ denotes the alpha value at the pixel location, while FG and BG denote the foreground and background image respectively. 
\let\saveeqnno\theequation
\let\savefrac\frac
\def\dispfrac{\displaystyle\savefrac}
\begin{eqnarray}
\let\frac\dispfrac
\gdef\theequation{1}
\let\theHequation\theequation
\label{dfg-3bd526b02979}
\begin{array}{@{}l}I_i\;=\;\alpha_i\times FG_i\;+\;(1-\alpha_i)\times BG_i\end{array}
\end{eqnarray}
\global\let\theequation\saveeqnno
\addtocounter{equation}{-1}\ignorespaces 
In the above-stated equation, only the RGB values of the images are known and the alpha values are still needed to be estimated. 

To simplify the estimation process of the alpha matte, many image matting models rely on user intervention in the form of a manually labeled trimap. A trimap is a densely labeled opacity map of the known and unknown regions in a given image. The known region includes the foreground and the background whose values are 1 and 0 respectively, whereas the unknown region includes the boundary regions whose opacity indices are not determined. The unknown regions are given an alpha value of 0.5. An alternative to trimap is stroke. Strokes are labeled regions of background and foreground that are manually marked using coarsely scribbled markings. As a result, stroke-based algorithms are lower in quality as a trade-off to less user input and faster inference speed. Another method for directly processing the RGB input to generate the matting features is spectral matting. Levin's spectral matting is a semi-automated approach that also requires user guidance to select the foreground features. The need for user intervention not only causes processing latency and expense in the matting workflow but also acutely limits the application of image matting.

Our proposal is a fully automated network for image segmentation-matting which can accurately segment out the target foreground from general natural images such as the COCO\unskip~\cite{632308:15988476} dataset. Our model takes an RGB image as an input and generates a highly accurate alpha matte for the target foreground, without additional user-labeled input. Our proposed deep learning approach encapsulates two stages in a single network, namely the segmentation phase which is responsible for generating a coarse trimap of the target foreground, followed by a matting phase which is responsible for converting the trimap prediction and the RGB image into a detailed alpha matte of the foreground objects. 

In the segmentation phase, an RGB image is fed into a DeepLabV3+ \unskip~\cite{632308:15988307} network, which predicts a binary mask of the target foreground. The predicted mask is used to retrieve the bounding box of the object, which is used as an informative input to estimate the trimap from the binary mask using an erosion-dilation (ED) layer at the end of the Deeplab network. Unlike other similar works \unskip~\cite{632308:15988477}, our ED module is built as a part of the network to make the system homogeneous in nature. A homogeneous network not only makes end-to-end training possible (Section 4.1) but also makes the system less prone to code library errors and faster in terms of inference speed. We chose semantic segmentation (returns the mask) over instance segmentation (returns the bounding box and the mask) because, on a CPU run-time, a semantic segmentation model gives a considerable advantage over an instance segmentation model in terms of inference speed, which is a very important factor that affects a production service. The trimap generation phase is followed by a matting phase, which uses the RGB image and the generated trimap to generate the final alpha matte. 

Our matting phase is inspired by the Deep Image Matting Network (DIM) \unskip~\cite{632308:15988513}. DIM largely borrows its inspiration from SegNet \unskip~\cite{632308:15988514}, which uses unpooling for upsampling. One of the major advantages of using SegNet for such a task is the recovery of boundary details, that are mostly missed by other architectures such as DeepLabv3+ and RefineNet \unskip~\cite{632308:15988515}. The major reason behind the fact that SegNet is better in recovering boundary details is that unpooling utilizes max-pooling's index guidance for upsampling. On the other hand, bi-linearly interpolated feature maps fail to emphasize on the boundary details. In order to record the boundary locations, the responses of the shallow layers of the network are used to project the excitation of various index locations on the feature maps into an attention mask.

In this paper, we not only demonstrate the effectiveness of AlphaNet on natural image matting but also on the quality of learned boundary details by visualizing learned indices, which shows that our network successfully captures the boundaries and textural patterns.
    
\section{Related Works}
Image matting methods for natural scenes has been under continuous research in the past few decades. Many of the developed methods predict the alpha matte through the propagation of color \unskip~\cite{632308:15988686,632308:15988687,632308:15988729,632308:15988730}, sampling \unskip~\cite{632308:15988516,632308:15988530,632308:15988531,632308:15988644} or low-level feature analysis. In the research literature of image matting, there are many non-learning types such as Bayesian-based \unskip~\cite{632308:15988516}, sampling-based\unskip~\cite{632308:15988731} and affinity-based methods like Poisson Matting \unskip~\cite{632308:15988732}, Closed-form Matting \unskip~\cite{632308:15988729} , and KNN Matting \unskip~\cite{632308:15988687}. However, in recent years, due to the rise of deep learning for computer vision, many convolutional neural network-based learning methods have been introduced for the task of general image matting in natural scenes. The advantage of learning-based methods is that the model learns the semantic meaning of the objects, thus performing exceptionally in cases where the background and foreground colors are similar. One such well-known network designs include deep image matting by Xu et al. \unskip~\cite{632308:15988513}. Nonetheless, all these methods require trimap or scribbles from the user as compulsion to estimate the alpha matte. Lately, types of research \unskip~\cite{632308:15988733},\unskip~\cite{632308:15988734} have proposed an automatic matting system. Zhu et al. \unskip~\cite{632308:15988734} and Shen et al. \unskip~\cite{632308:15988733} use fast filter similar to guided filters \unskip~\cite{632308:15988744} and closed-form matting with CNN respectively to automatically estimate the alpha mattes of portrait images. Chen et al. \unskip~\cite{632308:15988786} also introduced an automatic method for human matting that takes an RGB image as input and primarily predicts the background, foreground and transition region using the three-class segmentation network. The result of the segmentation phase is then used as a trimap for the alpha matte generation. Hu et el. \unskip~\cite{632308:15988477} proposed an integration of instance segmentation and image matting processes to generate alpha mattes. Zhang et el. \unskip~\cite{632308:15988828} proposes a convolutional network with two decoder branches for the foreground and background segmentation respectively, followed by a fusion branch that integrates the two classification results into an alpha matte. Aksoy et el. \unskip~\cite{632308:15988829} proposes a method for soft segmentation that captures the soft transitions between semantically meaningful regions by fusing high-level and low-level features of the image in a single graph structure. However, all the above-mentioned methods fail to recover detailed alpha mattes in case of high depth images, where the disparity between the foreground and the background is minimum.

Our other related work field is upsampling. Upsampling is an important stage in the network decoding phase for the task of dense prediction. The initial operator of upsampling was deconvolution which was initially used for visualizing the convolutional activations and was later extended to semantic segmentation. But deconvolution can easily have uneven overlap, putting more of the metaphorical paint in some places than others which leads to checkerboard artifacts. To avoid this uneven overlap of artifacts, resize + convolution paradigm was suggested. This paradigm presently serves as a standard configuration for most of the state-of-the-art semantic segmentation networks \unskip~\cite{632308:15988307},\unskip~\cite{632308:15988515}. Apart from these, unpooling \unskip~\cite{632308:15988514} and perforate \unskip~\cite{632308:15988830} operators were also suggested for generating sparse index maps to guide upsampling. However, these operators induce sparsity into the upsampled output but preserve the boundary information. \unskip~\cite{632308:15988831} introduces periodic shuffling which is a memory efficient and fast upsampling operator, generally used for the task of super-resolution. Lu et el. \unskip~\cite{632308:15988832} introduced a concept of index learning in their research through a novel index-guided encoder-decoder framework which use the data to automatically learn the indices to guide upsampling and pooling operator. 

In our work, we propose an attention module for the matting network, which can be considered as a combination of holistic and depthwise index networks \unskip~\cite{632308:15988832}. We realize the importance of spatial information, thus we preserve it during the encoding phase of the network. The stored spatial information is utilized by the encoding and decoding phase of the network during downsampling and upsampling respectively. For achieving the same, we use an attention module that self-learns the attention map from the data itself.
    
\section{Our Method}
In this section, we introduce the model architecture of the proposed method (Figure 2) and explain how each phase of the pipeline works individually and together. The model takes an RBG image as an input into the segmentation network and generate a binary segmentation mask for the foreground objects. The binary mask is used to estimate the bounding box, which is used as an input to the Erosion-Dilation (ED) Layer along with the mask to generate a trimap. The trimap generated by this process is coarse and contains many uncertain regions, mostly along the edges of the generated mask. This trimap is then concatenated with the RGB image and serves as an input to the matting network. The matting network is an attention guided model that estimates an alpha matte from the RGB image and the generated coarse trimap. The predicted alpha matte is then compared to the ground truth using different loss functions, and the gradient is calculated for network parameter optimization.

\bgroup
\fixFloatSize{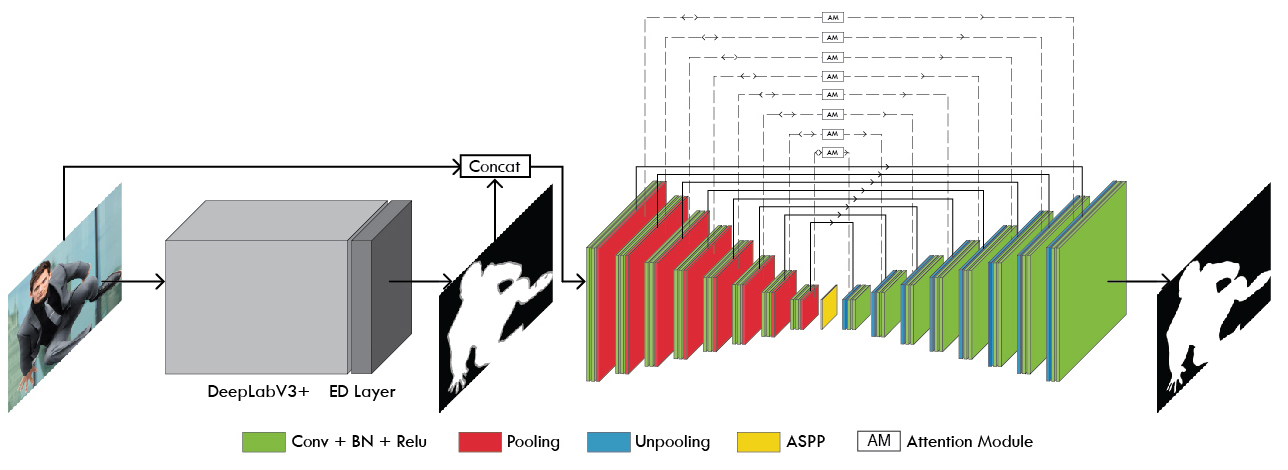}
\begin{figure*}[!htbp]
\centering \makeatletter\IfFileExists{7fb34c2a-a870-446c-9441-729aedef05a8-upaper3-01.jpg}{\includegraphics{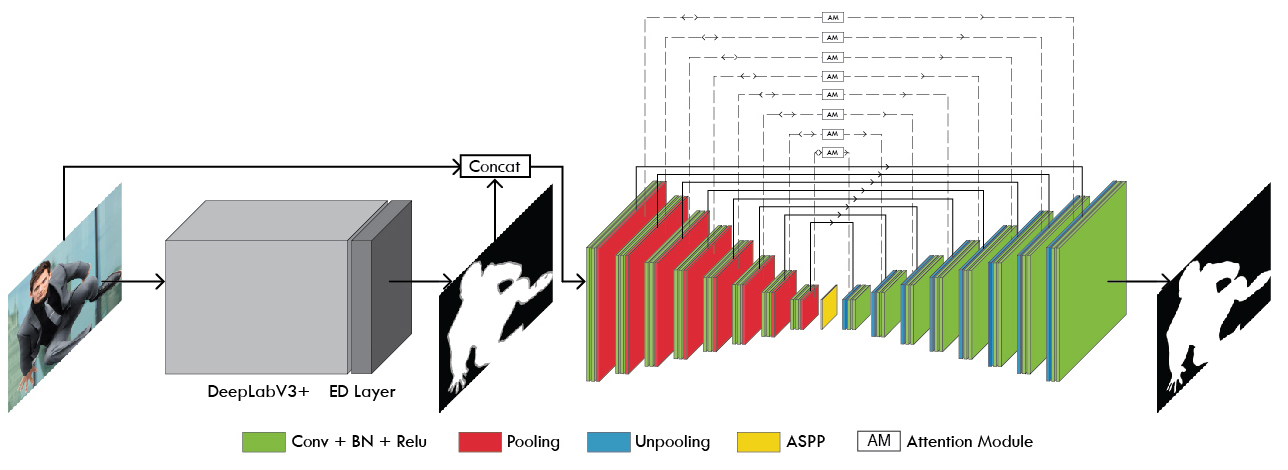}}{}
\makeatother 
\caption{{Overview of AlphaNet Architecture. Given an input image, a Segmentation-Trimap net, which is implemented as DeeplabV3+ with a ED Layer, is used to predict a trimap. The predicted trimap is then concatenated behind the input image as the fourth channel and fed into the matting net to predict the raw alpha matte which is processed with predicted trimap to generate the final alpha matte. The entire network is trained in an end-to-end fashion.}}
\label{f-875c0a9f8ff1}
\end{figure*}
\egroup

\subsection{Segmentation and Trimap Estimation Phase}The initial phase of AlphaNet is segmentation. The segmentation network comprises DeepLabV3+ \unskip~\cite{632308:15988307} encoder-decoder architecture with an additional erosion-dilation layer at the end to convert the binary output into a coarse trimap. The segmentation stage of our model could use any semantic segmentation model that produces a binary mask. We use DeepLabV3+ with a ReseNet18 backbone and pre-trained on the supervisely dataset\unskip~\cite{632308:16660404}. The pre-training dataset consists of 5711 images with 6884 high-quality annotated person instances. There are some additional advantages of pre-training on the supervisely dataset as compared to other public datasets. Human segmentation is a critical task in case of natural scenes, therefore a network pre-trained on such high-quality segmentation masks creates a robust initial checkpoint for training. 

Trimap estimation takes the output of the segmentation model along with additional derived information of the object bounding box. The trimap estimation phase makes an assumption that only the region near the mask boundary requires further estimation by the image matting model. This assumption is a major reason for the bottlenecks listed in Section 4.2.1 . A certain region eroded and dilated from the object binary mask is marked to be the unknown region in the trimap with $\alpha_i=0.5 $ . Other pixels that are inside the mask are classified as foreground with $\alpha_i=1.0 $ , while the pixels other than the unknown and foreground pixels are assigned $\alpha_i=0.0 $ . The extent of erosion and dilation is determined by the calculated object dimensions. Height is approximated by $height=bbox\lbrack3\rbrack\;-\;bbox\lbrack1\rbrack $ , whereas width is approximated by $width=bbox\lbrack2\rbrack\;-\;bbox\lbrack0\rbrack $ . The erosion and dilation rates are accordingly fixed as a percentage of the average height and width. However, there is a trade-off in adopting a lower or higher erosion-dilation rate. In our observation, when the true object boundary and the mask boundary are near, a small erosion-dilation rate is sufficient to cover the refinement region. But in some cases, when an object mask has errors along the boundaries, a large erosion-dilation rate tends to recover better and larger uncertainty region in the trimap. In order to estimate a precise alpha matte, a precise trimap is favored to impose a strong constraint on the matting half of the network.

\subsection{Matting Phase}Our matting phase is inspired by the learning-based method, proposed by Lin et al. \unskip~\cite{632308:15988513}, named as Deep Image Matting (DIM). DIM is a VGG16 based encoder-decoder network with an additional fully-connected refinement block. DIM shows exceptional results on images with a natural background where the color similarity between the foreground and background is very high. Unlike many non-learning methods, the quality of the trimap doesn't carry much dependency on the performance of this data-driven model. DIM was originally trained on 431 unique alpha mattes but the variety of the included objects is still narrow as compared to other known object classification datasets.

\subsubsection{Encoder-Decoder}Unlike Deep Image Matting, in our work, we built a MobileNetV2 \unskip~\cite{632308:15988874} based encoder-decoder with an additional attention module to guide the upsampling and downsampling operators, which compensates the utility of the refinement block in DIM. Using mobilenetv2 gives us an inference-time speedup on a CPU as compared to other heavy backbones. Figure 2 shows a basic overview of our network architecture. AlphaNet's matting phase follows a simple encoder-decoder paradigm with an attention module attached to all the pooling and unpooling layers. The pooling and unpooling layers follow a common configuration of 2 x 2 kernel size and 2 stride. At the core of our network is the attention module, that ingests the feature maps from the encoder branch and generates an attention map to guide the downsampling and upsampling operators. In this research, we also investigate alternate ways for context encoding and low-level feature fusion, but due to lack of any substantial-conclusion, we skip sharing the metrics for the same.

\subsubsection{Attention Module}Our attention module models the attention map as a function of the encoder feature map $F\in\mathbb{R}^{H\;x\;W\;x\;C} $. The module generates two attention maps for the upsampling and downsampling operators respectively. The attention map has the same spatial dimension as the input feature map but with just one channel containing a specific attention weightage $A_i $ for every index i in F, $A_i\in\lbrack0,1\rbrack $ . The chosen range for $A_i $ provides smooth optimization and helps in better convergence. The attention mechanics consists of a predefined attention block followed by two normalization layers. The core of the attention block consists of a fully convolutional neural network that interprets an input feature map into an attention map. The attention block learns an attention function .
\begin{eqnarray*}A(F):\;\mathbb{R}^{H\;x\;W\;x\;C}\xrightarrow{}\mathbb{R}^{H\;x\;W\;x\;1} \end{eqnarray*}
 As shown in the attention function, all channels of the feature map get projected into a single attention map. The attention block is followed by two normalization layers that are responsible for normalizing the attention map for encoder and decoder differently. The attention map for the encoder is first normalized by a sigmoid function, which is followed by another normalization by a softmax function. The two normalization of the encoder attention map ensure the magnitude consistency of the feature map post downsampling. However, the decoder attention map only gets normalized by a sigmoid function. Once the attention maps are normalized, they are fed to the encoder and decoder pooling and unpooling operators respectively. In the case of an encoder, given a local region $L^{E}\in\mathbb{R}^{k\;x\;k} $ and its computed attention map $A^{L^{E}}=A(L^{E}) $, we compute a sum pooling operation of the element-wise multiplication $L^{E}\otimes A^{L^{E}} $. However, in the case of decoder, given a local region $L^{D}\in\mathbb{R}^{k\;x\;k} $ and its computed attention map $A^{L^{D}}=A(L^{D}) $, we upsample the element-wise multiplication $L^{D}\otimes A^{L^{D}} $. The major difference between normal pooling-unpooling and our method is that normal operation applies a fixed learned kernel to all the regions, whereas our module applies different kernels to different regions, based on the calculated attention maps. Concrete design of the attention module can be seen in Figure 3. The attention module assumes a non-linear relationship between the attention map and the feature map. The assumption is natural because a linear function cannot even fit the max function, so assuming a non-linear relationship makes the network more flexible in terms of learning abstract relationships inside the feature maps. The module is implemented as a fully convolutional network. It first uses four parallel 4 x 4 group convolution with 2-stride, 1 padding, and 2 groups on a feature map of dimensions H x W x C, generating an attention map of size H/2 x W/2 x 2C. This operation is followed by a group normalization \unskip~\cite{632308:15988876} layer and a ReLU activation for nonlinear mappings. The generated tensor is then processed with two point-wise convolution layer to achieve feature map pooling, generating an attention map of dimension H/2 x W/2 x 1. The final attention map is composed of the four downsampled attention maps by shuffling and rearrangement (Pixel Shuffle upsampling). Note that we experimented a lot with different network configurations for the attention module, and most of them achieved similar results with slight improvements. It's also important to note that the parameters of four convolutional layers in the attention module are not shared.

\bgroup
\fixFloatSize{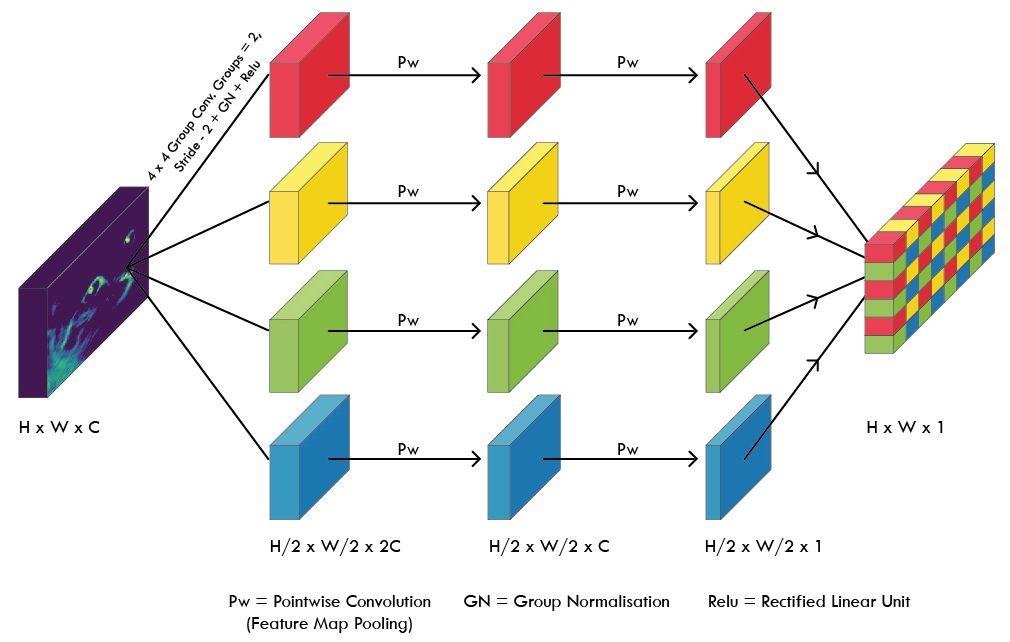}
\begin{figure}[t!]
\centering \makeatletter\IfFileExists{3c48681b-e6c5-4a06-9fea-5d39ffb5225e-upaper3-02.jpg}{\includegraphics{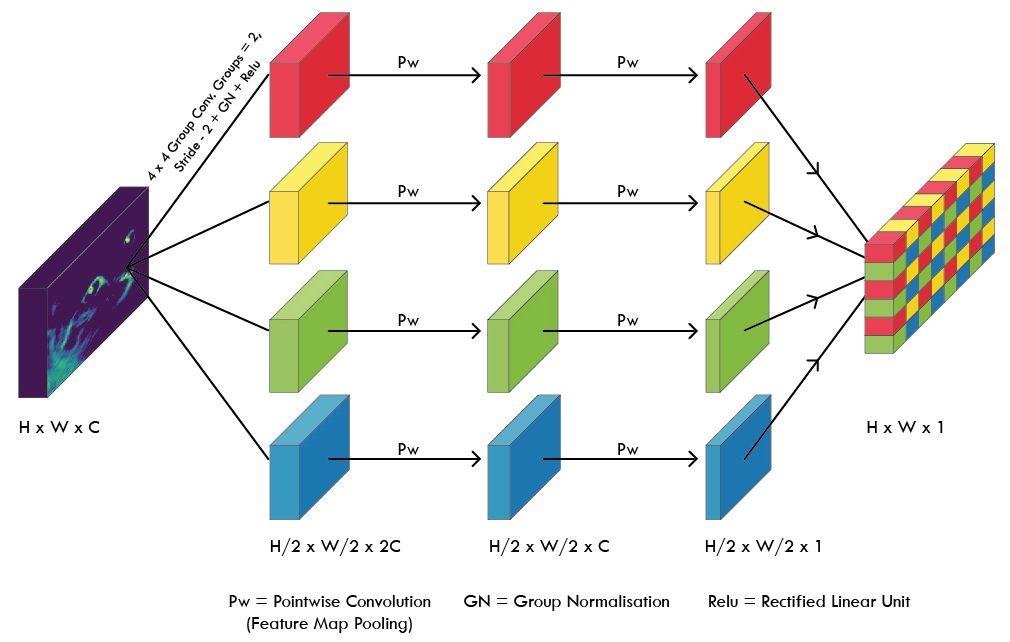}}{}
\makeatother 
\caption{{Attention Module}}
\label{f-5a1add349f45}
\end{figure}
\egroup

\subsection{Relation to Other Networks}In this subsection, we discuss our relationship with some of the published researches that have a similar spirit.

\textbf{Indices Matter (IndexNet) \unskip~\cite{632308:15988832}}. IndexNet suggests a plug-in indexing module applying to any off-the-shelf CNN, using two major types of index networks, DIN and HIN. Both HIN and DIN have their pros and cons. One of the major pros of IndexNet is its flexible network module which learns indices adaptively from the data to guide the downsampling and upsampling operators. However, its cons include the overfitting tendency of DIN and the lower capacity of HIN. Also, due to the use of batch normalization in the index module, the parameter tuning on small batch sizes becomes difficult. On the other hand, the attention module of AlphaNet uses group normalization and different network architecture, that make use of the better half of HIN and DIN, while ruling out the negative sides.

\textbf{Attention Networks \unskip~\cite{632308:15988918}}. AlphaNet shares a very close resemblance with the present existing family of attention networks, that works on a simple mechanism of multiplicative operation between a generated attention map and the feature map. Unlike other attention networks that work on refining the feature map, our attention mechanism just emphasize on guiding the upsampling and downsampling operators.

\textbf{Deformable Convolution Networks (DCNs) \unskip~\cite{632308:15988919}. }DCN proposes deformable convolution and RoI pooling by predicting the offsets of convolutional and pooling kernels respectively. DCN also seeks to enhance the transformation modeling capability of convolutional networks by giving a dynamic nature to CNN. AlphaNet also shares the same spirit of making a dynamic learning module that learns an attention map from the data itself.

\bgroup
\fixFloatSize{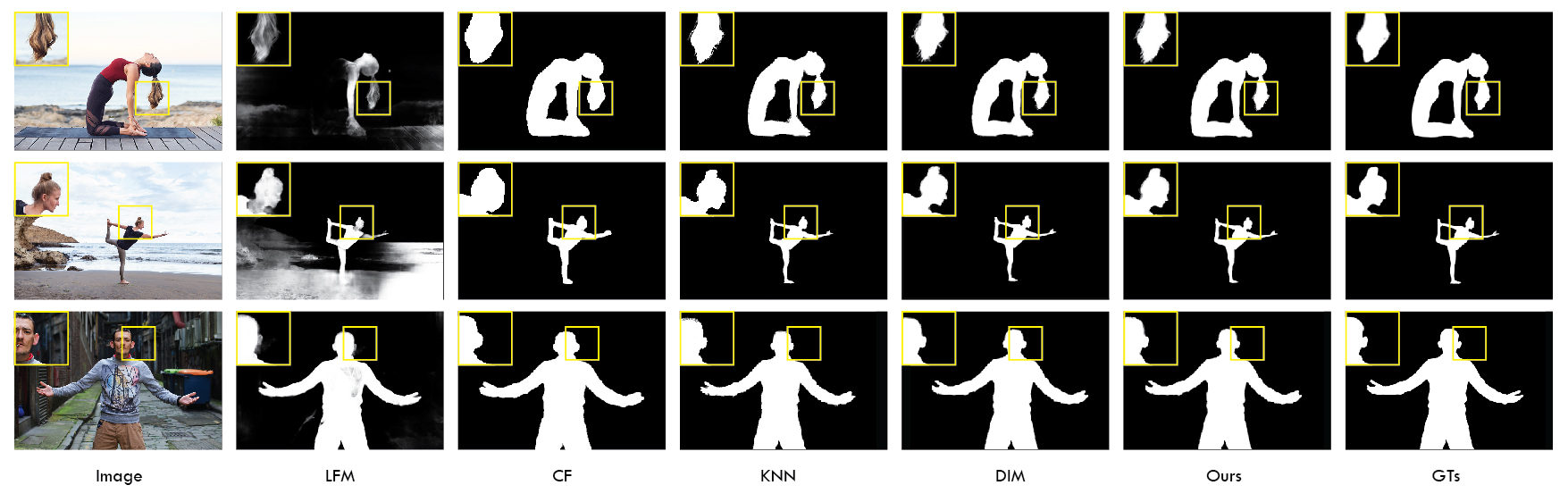}
\begin{figure*}[!htbp]
\centering \makeatletter\IfFileExists{d94bd4bc-b012-4480-9480-c3605c8acdc7-ucollage-01.jpg}{\includegraphics{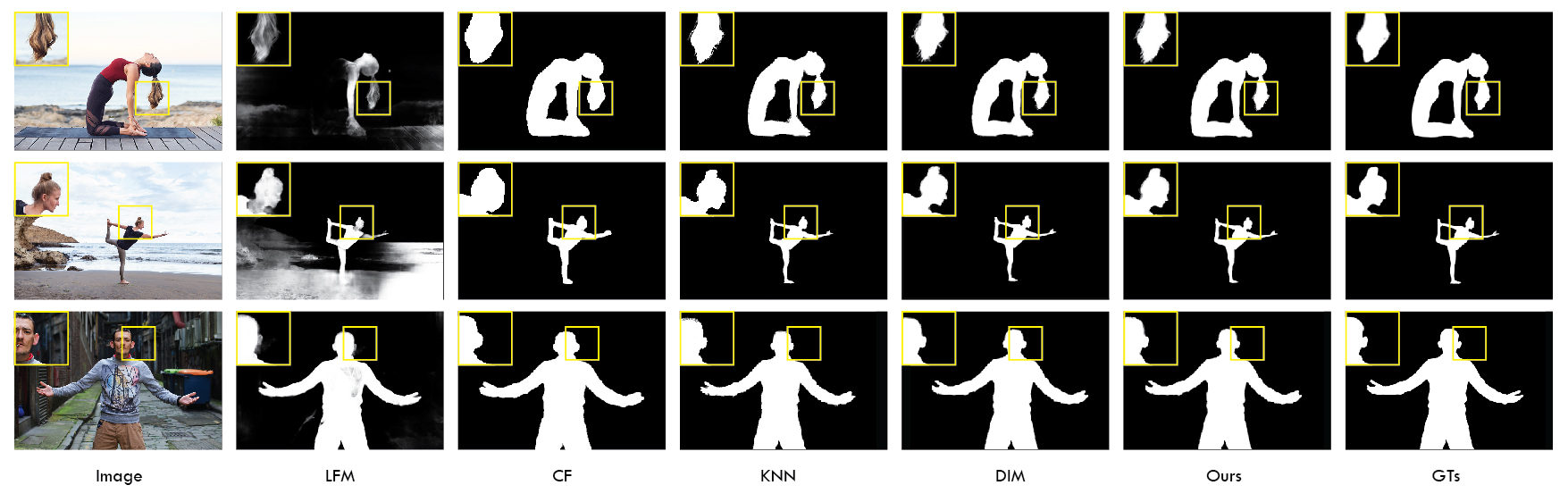}}{}
\makeatother 
\caption{{Qualitative results on the testing set. From left to right, the original image, LFM(pre-trained on DIM), closed-form matting, KNN matting, DIM with Refinement block, AlphaNet and ground-truth alpha matte.}}
\label{f-3376c4f36388}
\end{figure*}
\egroup

\section{Results and Discussion}
We implement our proposed network with PyTorch \unskip~\cite{632308:15989027}, a framework for coding deep networks. The segmentation and matting stages are first pre-trained and then fine-tuned end to end as explained in this section.

\begin{table}[!htbp]
\caption{{Quantitative Results on the test Split. Top results are emphasized in bold.} }
\label{tw-aed2fe692c4b}
\def\arraystretch{1}
\ignorespaces 
\centering 
\begin{tabulary}{\linewidth}{p{\dimexpr.3342\linewidth-2\tabcolsep}p{\dimexpr.15029999999999998\linewidth-2\tabcolsep}p{\dimexpr.17089999999999996\linewidth-2\tabcolsep}p{\dimexpr.15970000000000004\linewidth-2\tabcolsep}p{\dimexpr.1849\linewidth-2\tabcolsep}}
\hline Methods & MSE ($\times10^{-3} $) & SAD ($\times10^{-3} $) & Gradient ($\times10^{-5} $) & Connectivity ($\times10^{-5} $)\\
\hline 
DeeplabV3+ + CF &
  5.132 &
  8.761 &
  21.528 &
  43.152\\
DeeplabV3+ + IFM &
  4.271 &
  7.566 &
  19.712 &
  52.414\\
DeeplabV3+ + KNN &
  4.315 &
  7.609 &
  20.416 &
  56.319\\
DeeplabV3+ + DIM &
  3.743 &
  6.002 &
  19.397 &
  41.748\\
LMF (pre-trained on DIM dataset) &
  5.725 &
  8.641 &
  22.172 &
  42.021\\
AlphaNet w/o Attention &
  3.941 &
  6.820 &
  20.143 &
  42.672\\
\cellcolor[HTML]{B3B3B3}{AlphaNet} &
  \cellcolor[HTML]{CCCCCC}{3.794} &
  \cellcolor[HTML]{CCCCCC}{6.159} &
  \cellcolor[HTML]{CCCCCC}{19.442} &
  \cellcolor[HTML]{CCCCCC}{42.001}\\
\hline 
\end{tabulary}\par 
\end{table}
\textbf{Dataset:} We trained our model on a self-curated large scale high-quality image matting dataset, emphasized majorly on human portraits, fashion accessories and some strongly, medium and little transparent objects (the classification of transparent objects can be seen in the test set of \unskip~\cite{632308:15988920} {\textemdash} highly Transparent, strongly transparent, medium transparent and little transparent). Our dataset has 25,236 training image and 1,500 testing images. It is important to note that only a fraction of the mentioned number of images is unique foregrounds, composed of different backgrounds from MS COCO \unskip~\cite{632308:15988476} to reach the mentioned number. We evaluate our method on a test set made from our own curated dataset and the Adobe Image Matting test set \unskip~\cite{632308:15988513}. It is also important to note that, our dataset has samples from sensitive commercial sources, because of which it cannot be open-sourced.

\textbf{Measurement}: We evaluated the quality of the output matte using four metrics: Mean Squared Error (MSE), Sum of Absolute Differences (SAD), and perceptually motivated Connectivity (Conn) and Gradient (Grad) errors. Mean squared error and sum of absolute differences are directly correlated to the training objective, whereas the connectivity and gradient error evaluate the perceptual visual quality as assessed by a human observer. To calculate these metrics, we normalized the ground truth and predicted alpha to a range between 0 and 1. Also, unlike other similar works, all our metrics were calculated over the entire image instead of the unknown regions, followed by a basic average by the number of pixels. The evaluation code implemented similar to \unskip~\cite{632308:15988513} is used. We also perform extensive ablation studies to justify choices of model design.

\textbf{Baselines and Comparative Models}

We evaluated the capability and effectiveness of AlphaNet by comparing it with some of the following state-of-the-art matting methods: KNN matting \unskip~\cite{632308:15988687}, Late Fusion Matting (pre-trained on the DIM dataset) \unskip~\cite{632308:15988828}, Deep Image Matting (DIM) \unskip~\cite{632308:15988513}, Close Form matting (CF) \unskip~\cite{632308:15988729} and Information Flow Matting (IFM) \unskip~\cite{632308:15988686}. Some of these matting methods are interactive in nature and require trimap as a fourth channel input alongside the RGB. Therefore, to achieve a fair evaluation and comparison, we connect them with the predicted trimaps of our segmentation and trimap generation phase (Section 3.1) and also evaluated them separately by providing the trimap ground truths.

\begin{table}[!htbp]
\caption{{Quantitative Results on the test Split with trimap GT for interactive models. Top results are emphasized in bold.} }
\label{tw-ea2652d8e68f}
\def\arraystretch{1}
\ignorespaces 
\centering 
\begin{tabulary}{\linewidth}{p{\dimexpr.3095\linewidth-2\tabcolsep}p{\dimexpr.16389999999999997\linewidth-2\tabcolsep}p{\dimexpr.1529\linewidth-2\tabcolsep}p{\dimexpr.18200000000000003\linewidth-2\tabcolsep}p{\dimexpr.1917\linewidth-2\tabcolsep}}
\hline Methods & MSE ($\times10^{-3} $) & SAD ($\times10^{-3} $) & Gradient ($\times10^{-5} $) & Connectivity ($\times10^{-5} $)\\
\hline 
TrimapGT + CF &
  4.942 &
  8.592 &
  21.149 &
  42.861\\
TrimapGT + IFM &
  4.155 &
  7.238 &
  19.860 &
  52.217\\
TrimapGT + KNN &
  3.952 &
  6.984 &
  19.539 &
  53.628\\
TrimapGT + DIM &
  3.612 &
  5.931 &
  19.179 &
  40.927\\
\cellcolor[HTML]{B3B3B3}{AlphaNet} &
  \cellcolor[HTML]{CCCCCC}{3.794} &
  \cellcolor[HTML]{CCCCCC}{6.159} &
  \cellcolor[HTML]{CCCCCC}{19.442} &
  \cellcolor[HTML]{CCCCCC}{42.001}\\
\hline 
\end{tabulary}\par 
\end{table}

\bgroup
\fixFloatSize{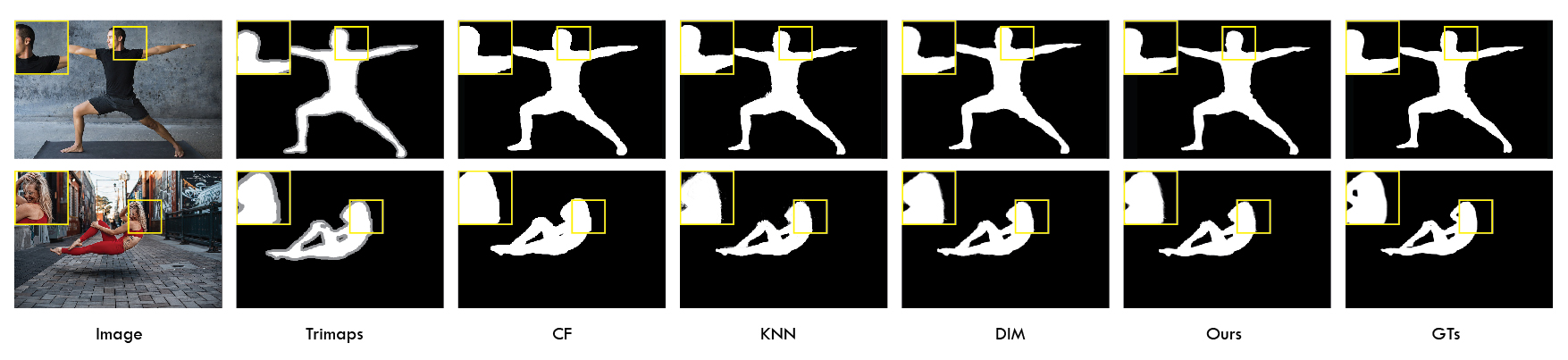}
\begin{figure*}[!htbp]
\centering \makeatletter\IfFileExists{aadcf477-e123-4bb4-9b88-dd13007e4620-ucollage-02.jpg}{\includegraphics{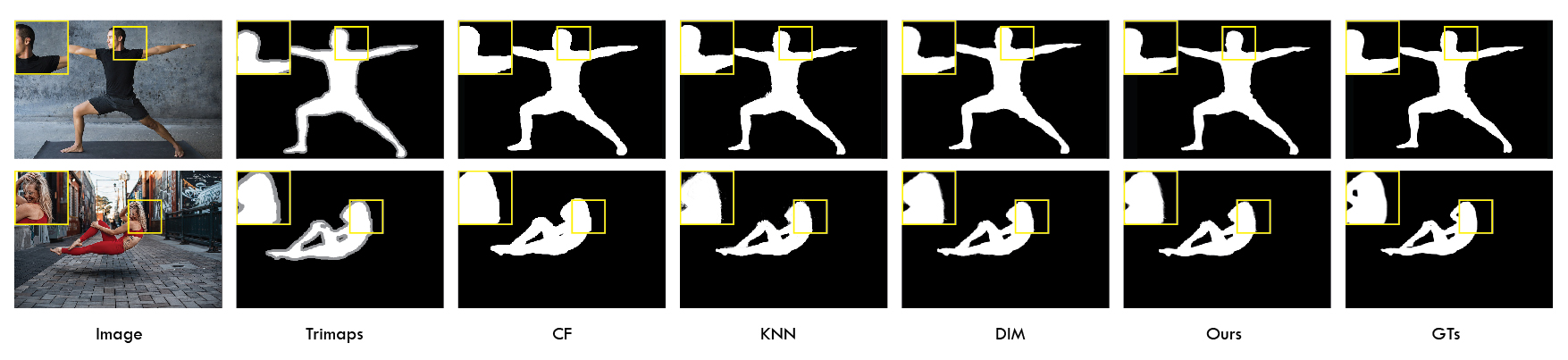}}{}
\makeatother 
\caption{{Qualitative results on the testing set with Trimaps. From left to right, the original image, Trimaps, Closed-form matting, KNN matting, DIM with Refinement block, AlphaNet and ground-truth alpha matte.}}
\label{f-1d0a54d34689}
\end{figure*}
\egroup

\subsection{Implementation Details}In this section, we describe some essential training details and other adopted strategies. We used the pre-train technique \unskip~\cite{632308:15988932}\unskip~\cite{632308:16646768}, a very effective and widely adopted technique in the deep learning domain. Following this practice, we first pre-train the segmentation and matting net separately and then the two nets are combined to make AlphaNet and finetuned in an end-to-end manner. One major advantage of adopting this technique is that we can use extra data specific to the sub-tasks for training the networks. Also, it was carefully noted that the pre-training set does not overlap with the testing set.

\textbf{Segmentation Net Pre-training}: Segmentation net comprise a DeeplabV3+ model with an additional erosion-dilation layer at the end. Since the ED layer is non-differentiable, the segmentation net was pre-trained as a normal semantic segmentation model. For training, images were uniformly resized to 496 x 496 spatial resolution, followed by the augmentation of the samples to avoid overfitting. It is worth noting that the model was pre-trained on supervisely dataset

\textbf{Matting Net Pre-training}: To pre-train the matting net, we follow a similar training configuration as used in deep image matting \unskip~\cite{632308:15988513}. For data augmentation, we used 320 x 320 random cropping, followed by random flipping and random scaling. We used ImageNet \unskip~\cite{632308:15989025} pre-trained parameters for the encoder as the initial checkpoint. For initializing all the other parameters, we utilized a better form of \unskip~\cite{632308:15989026}, which was proposed in \unskip~\cite{632308:15988933}. \unskip~\cite{632308:15988933} shows that "Xavier" has a very direct conclusion for the derived variance being adopted for Gaussian distributions only. We used Adam \unskip~\cite{632308:15988940} for optimization and trained the net for about 50 epochs. The batch size is set to 16 along with fixing the BN layers of the backbone.

\textbf{AlphaNet Training}: In order to perform end-to-end training of AlphaNet, we first initialize AlphaNet with the pre-trained segmentation and matting net. During training, the input samples are augmented on-the-fly using random horizontal flipping (0.5 probability). We also rescale the longer dimension of the input if it exceeds 1500 pixels to cap it to 1500 for the GPU memory limitation. The predicted alpha matte is then rescaled back to its original size for the evaluation. This technique enables AlphaNet to perform testing on CPU for high-resolution images, without any significant loss of resolution. One major catch in the end-to-end training is that the gradient is unable to flow from the matting net to the segmentation net, due to the non-parametric nature of the erosion-dilation layer. Thus, only the loss value of the matting net flows back as a feedback to the segmentation net (end-to-end non-differentiable).

\textbf{Loss}: For pre-training the segmentation net, we mainly used a combination focal \unskip~\cite{632308:15988941} and dice loss. For pre-training the matting net, we adopted compositional and alpha prediction loss. The compositional loss ( $C_L $) can be defined as the absolute difference between the predicted alpha compositional image and the ground truth compositional image. Whereas, the alpha prediction loss ( $A_L $) can be defined as the absolute difference between the predicted alpha and the ground truth alpha. The overall loss is then calculated according to the following equation.
\begin{eqnarray*}Loss\;=\;\gamma A_L\;+\;(1-\gamma)C_L \end{eqnarray*}
Where $A_L\;=\;\parallel\;\alpha_G\;-\;\alpha_P\parallel $ , $C_L\;=\;\parallel\;c_G\;-\;c_P\parallel $ and $\gamma\;=\;0.5 $  in our case. Unlike similar works that only compute loss over the unknown regions, our prediction loss is summed over the complete image.

\subsection{Performance Comparison}In this sub-section, we study the comparison of AlphaNet vs the state-of-the-art image matting models. To get a fair comparison, we feed the interactive models with our segmentation net generated trimaps. Table 1 summarizes the quantitative result of our evaluation.

Considering the complex structural details to be learned by the models, the resulting metric shows that the automatic models outperform most of the interactive matting models. DIM connected with DeeplabV3+ outperforms all the other models, mainly due to its efficiency to learn complex context and the add-on advantage of the extra refinement block. But we can see from the results that AlphaNet not only performs better than most of the baselines but also better than LFM (a completely automatic method). The major reason is the attention learning capability of our model that enables it to learn fine details and coarse semantics. Fig. 4 summarizes several visual evaluations and respective comparisons with all the baselines. It can be clearly seen from the results that our model was able to recover subtle details and generate much clearer mattes with good local contrast.

\subsubsection{Dataset Bottleneck}There are majorly four categories of data in the image matting datasets. The categories include highly transparent, strongly transparent, medium transparent and little transparent entities \unskip~\cite{632308:15988920}. One major drawback of our work is that we emphasized our dataset on only strongly, medium and little transparent objects. We dropped the category of highly transparent objects because of the low availability of such data in context with our fashion e-commerce industry. Therefore, as our network was not exposed to highly transparent objects, it fails to model the transparency of pixels deep within the boundaries of the object. We plan to remove this data bottleneck in our next version and come up with a more generic model that covers the matting of highly transparent objects and also extends our dataset domain wider than just fashion e-commerce.

\subsection{Ablative Study of Different Components}\textbf{Effectiveness of End-to-end Training}

The end-to-end training strategy's effectiveness is calculated by comparing end-to-end trained AlphaNet with the one consisting of only pre-trained parameters. Table 1 shows the results for the same. It can be easily observed from the result metric that end-to-end trained AlphaNet outperforms individually pre-trained nets, thus proving the effectiveness of end-to-end training. 

\textbf{The Evaluation of Attention Module}

The effectiveness and importance of our proposed attention module is validated by simply comparing the version of AlphaNet with an attention module to another baseline version with no attention module. The attention-less module is also trained for the same objective as the attention version. On comparing the evaluation metrics of the two versions, it can be seen that AlphaNet with an attention module performs better than the baseline version. It is also worth noting that although most of the metrics are quantitatively small, connectivity and gradient error of the non-attention baseline is relatively large. The major reason for the large error is the blurring of structural details, induced by the encoder-decoder without external guidance of an attention network. Therefore, our proposed attention module not only leverages the coarse estimation of the generated trimap, but also plays a key role in enhancing the overall performance of AlphaNet.
    
\section{Conclusion}
In this work, we propose AlphaNet, a fully automated image matting method. Our method does not require any human interaction to predict smooth alpha mattes. The proposed method can be seen both as a fully automated semantic image matting method and a refinement of existing semantic segmentation models that work on common datasets. We compared and presented our results with the related works and proved how AlphaNet not only outperforms many but also shows comparable performance with the remaining. We also proposed a novel attention module that along with the encoder-decoder net, captures local details and global semantic context to generate high-quality alpha mattes automatically. Furthermore, we also created a high-quality fashion e-commerce oriented dataset. Benefiting from our own curated dataset and model architecture, our method registers comparable results with state-of-the-art matting methods.

\section*{Acknowledgments}We would like to thank our organization for the constant support, resources and valuable suggestions for conducting this research. 


%

\bibliographystyle{IEEEtran}

\bibliography{\jobname}

\begin{thebibliography}{10}
\providecommand{\url}[1]{#1}
\csname url@samestyle\endcsname
\providecommand{\newblock}{\relax}
\providecommand{\bibinfo}[2]{#2}
\providecommand{\BIBentrySTDinterwordspacing}{\spaceskip=0pt\relax}
\providecommand{\BIBentryALTinterwordstretchfactor}{4}
\providecommand{\BIBentryALTinterwordspacing}{\spaceskip=\fontdimen2\font plus
\BIBentryALTinterwordstretchfactor\fontdimen3\font minus
  \fontdimen4\font\relax}
\providecommand{\BIBforeignlanguage}[2]{{%
\expandafter\ifx\csname l@#1\endcsname\relax
\typeout{** WARNING: IEEEtran.bst: No hyphenation pattern has been}%
\typeout{** loaded for the language `#1'. Using the pattern for}%
\typeout{** the default language instead.}%
\else
\language=\csname l@#1\endcsname
\fi
#2}}
\providecommand{\BIBdecl}{\relax}
\BIBdecl

\bibitem{632308:15988476}
T.-Y. Lin, M.~Maire, S.~J. Belongie, L.~D. Bourdev, R.~B. Girshick, J.~Hays,
  P.~Perona, D.~Ramanan, P.~Doll\'{a}r, and C.~L. Zitnick, ``{Microsoft COCO:
  Common Objects in Context},'' \emph{{CoRR}}, vol. abs/1405.0312, 2014.

\bibitem{632308:15988307}
L.-C. Chen, Y.~Zhu, G.~Papandreou, F.~Schroff, and H.~Adam, ``{Encoder-Decoder
  with Atrous Separable Convolution for Semantic Image Segmentation},''
  \emph{{CoRR}}, vol. abs/1802.02611, 2018.

\bibitem{632308:15988477}
G.~Hu and J.~J. Clark, ``{Instance Segmentation based Semantic Matting for
  Compositing Applications},'' \emph{{CoRR}}, vol. abs/1904.05457, 2019.

\bibitem{632308:15988513}
N.~Xu, B.~L. Price, S.~Cohen, and T.~S. Huang, ``{Deep Image Matting},''
  \emph{{CoRR}}, vol. abs/1703.03872, 2017.

\bibitem{632308:15988514}
V.~Badrinarayanan, A.~Kendall, and R.~Cipolla, ``{SegNet: A Deep Convolutional
  Encoder-Decoder Architecture for Image Segmentation},'' \emph{{IEEE
  Transactions on Pattern Analysis and Machine Intelligence}}, vol.~39, pp.
  2481--2495, 2017.

\bibitem{632308:15988515}
G.~Lin, A.~Milan, C.~Shen, and I.~D. Reid, ``{RefineNet: Multi-Path Refinement
  Networks for High-Resolution Semantic Segmentation},'' \emph{{CoRR}}, vol.
  abs/1611.06612, 2016.

\bibitem{632308:15988686}
Y.~Aksoy, T.~O. Aydin, and M.~Pollefeys, ``{Designing Effective Inter-Pixel
  Information Flow for Natural Image Matting},'' \emph{{CoRR}}, vol.
  abs/1707.05055, 2017.

\bibitem{632308:15988687}
Q.~Chen, D.~Li, and C.-K. Tang, ``{KNN Matting},'' \emph{{IEEE Transactions on
  Pattern Analysis and Machine Intelligence}}, vol.~35, pp. 2175--2188, 2012.

\bibitem{632308:15988729}
A.~Levin, D.~Lischinski, and Y.~Weiss, ``{A Closed-Form Solution to Natural
  Image Matting},'' \emph{{IEEE Transactions on Pattern Analysis and Machine
  Intelligence}}, vol.~30, pp. 228--242, 2006.

\bibitem{632308:15988730}
L.~Grady and R.~Westermann, ``{RANDOM WALKS FOR INTERACTIVE ALPHA-MATTING},''
  2005.

\bibitem{632308:15988516}
``{A Bayesian Approach to Digital Matting},'' in \emph{{Proceedings of IEEE
  CVPR 2001}}, vol.~2.\hskip 1em plus 0.5em minus 0.4em\relax IEEE Computer
  Society, December 2001, pp. 264--271.

\bibitem{632308:15988530}
K.~He, C.~Rhemann, C.~Rother, X.~Tang, and J.~Sun, ``{A global sampling method
  for alpha matting},'' \emph{{CVPR 2011}}, pp. 2049--2056, 2011.

\bibitem{632308:15988531}
E.~Shahrian, D.~Rajan, B.~L. Price, and S.~Cohen, ``{Improving Image Matting
  Using Comprehensive Sampling Sets},'' \emph{{2013 IEEE Conference on Computer
  Vision and Pattern Recognition}}, pp. 636--643, 2013.

\bibitem{632308:15988644}
J.~Wang and M.~F. Cohen, ``{Optimized Color Sampling for Robust Matting},''
  \emph{{2007 IEEE Conference on Computer Vision and Pattern Recognition}}, pp.
  1--8, 2007.

\bibitem{632308:15988731}
M.~A. Ruzon and C.~Tomasi, ``{Alpha estimation in natural images},''
  \emph{{Proceedings IEEE Conference on Computer Vision and Pattern
  Recognition. CVPR 2000 (Cat. No.PR00662)}}, vol.~1, 2000.

\bibitem{632308:15988732}
E.~S.~L. Gastal and M.~M. de~Oliveira~Neto, ``{Shared Sampling for Real-Time
  Alpha Matting},'' \emph{{Comput. Graph. Forum}}, vol.~29, pp. 575--584, 2010.

\bibitem{632308:15988733}
X.~Shen, X.~Tao, H.~Gao, C.~Zhou, and J.~Jia, ``{Deep Automatic Portrait
  Matting},'' in \emph{{ECCV}}, 2016.

\bibitem{632308:15988734}
B.~Zhu, Y.~Chen, J.~Wang, S.~Liu, B.~Zhang, and M.~Tang, ``{Fast Deep Matting
  for Portrait Animation on Mobile Phone},'' \emph{{CoRR}}, vol.
  abs/1707.08289, 2017.

\bibitem{632308:15988744}
\BIBentryALTinterwordspacing
K.~He, J.~Sun, and X.~Tang, ``{Guided Image Filtering},'' \emph{{IEEE
  transactions on pattern analysis and machine intelligence}}, vol.~35, pp.
  1397--1409, 06 2013. [Online]. Available: \url{10.1109/TPAMI.2012.213}
\BIBentrySTDinterwordspacing

\bibitem{632308:15988786}
Q.~Chen, T.~Ge, Y.~Xu, Z.~Zhang, X.~Yang, and K.~Gai, ``{Semantic Human
  Matting},'' \emph{{CoRR}}, vol. abs/1809.01354, 2018.

\bibitem{632308:15988828}
Y.~Zhang, L.~Gong, L.~Fan, P.~Ren, Q.~Huang, H.~Bao, and W.~Xu, ``{A Late
  Fusion CNN for Digital Matting},'' in \emph{{The IEEE Conference on Computer
  Vision and Pattern Recognition (CVPR)}}, June 2019.

\bibitem{632308:15988829}
Y.~Aksoy, T.-H. Oh, S.~Paris, M.~Pollefeys, and W.~Matusik, ``{Semantic Soft
  Segmentation},'' \emph{{ACM Transactions on Graphics (Proc. SIGGRAPH)}},
  vol.~37, no.~4, 2018.

\bibitem{632308:15988830}
C.~Osendorfer, H.~Soyer, and P.~van~der Smagt, ``{Image Super-Resolution with
  Fast Approximate Convolutional Sparse Coding},'' in \emph{{ICONIP}}, 2014.

\bibitem{632308:15988831}
W.~Shi, J.~Caballero, F.~Husz\'{a}r, J.~Totz, A.~P. Aitken, R.~Bishop,
  D.~Rueckert, and Z.~Wang, ``{Real-Time Single Image and Video
  Super-Resolution Using an Efficient Sub-Pixel Convolutional Neural
  Network},'' \emph{{CoRR}}, vol. abs/1609.05158, 2016.

\bibitem{632308:15988832}
H.~Lu, Y.~Dai, C.~Shen, and S.~Xu, ``{Indices Matter: Learning to Index for
  Deep Image Matting},'' in \emph{{Proc. IEEE/CVF International Conference on
  Computer Vision (ICCV)}}, 2019.

\bibitem{632308:16660404}
\BIBentryALTinterwordspacing
 [Online]. Available: \url{https://www.supervise.ly/}
\BIBentrySTDinterwordspacing

\bibitem{632308:15988874}
M.~Sandler, A.~G. Howard, M.~Zhu, A.~Zhmoginov, and L.-C. Chen, ``{Inverted
  Residuals and Linear Bottlenecks: Mobile Networks for Classification,
  Detection and Segmentation},'' \emph{{CoRR}}, vol. abs/1801.04381, 2018.

\bibitem{632308:15988876}
Y.~Wu and K.~He, ``{Group Normalization},'' \emph{{CoRR}}, vol. abs/1803.08494,
  2018.

\bibitem{632308:15988918}
V.~Mnih, N.~Heess, A.~Graves, and K.~Kavukcuoglu, ``{Recurrent Models of Visual
  Attention},'' \emph{{CoRR}}, vol. abs/1406.6247, 2014.

\bibitem{632308:15988919}
J.~Dai, H.~Qi, Y.~Xiong, Y.~Li, G.~Zhang, H.~Hu, and Y.~Wei, ``{Deformable
  Convolutional Networks},'' \emph{{CoRR}}, vol. abs/1703.06211, 2017.

\bibitem{632308:15989027}
A.~Paszke, S.~Gross, S.~Chintala, G.~Chanan, E.~Yang, Z.~DeVito, Z.~Lin,
  A.~Desmaison, L.~Antiga, and A.~Lerer, ``{Automatic differentiation in
  PyTorch},'' 2017.

\bibitem{632308:15988920}
C.~Rhemann, C.~Rother, J.~Wang, M.~Gelautz, P.~Kohli, and P.~Rott, ``{A
  Perceptually Motivated Online Benchmark for Image Matting},'' in
  \emph{{Proceddings of the IEEE Conference on Computer Vision and Pattern
  Recognition}}, 2009.

\bibitem{632308:15988932}
\BIBentryALTinterwordspacing
G.~Hinton, S.~Osindero, and Y.-W. Teh, ``{A Fast Learning Algorithm for Deep
  Belief Nets},'' \emph{{Neural computation}}, vol.~18, pp. 1527--54, 08 2006.
  [Online]. Available: \url{10.1162/neco.2006.18.7.1527}
\BIBentrySTDinterwordspacing

\bibitem{632308:16646768}
R.~Sharma and A.~Vishvakarma, ``{Retrieving Similar E-Commerce Images Using
  Deep Learning},'' \emph{{CoRR}}, vol. abs/1901.03546, 2019.

\bibitem{632308:15989025}
``{ImageNet: A Large-Scale Hierarchical Image Database},'' in \emph{{CVPR09}},
  2009.

\bibitem{632308:15989026}
X.~Glorot and Y.~Bengio, ``{Understanding the difficulty of training deep
  feedforward neural networks},'' \emph{{Journal of Machine Learning Research -
  Proceedings Track}}, vol.~9, pp. 249--256, 01 2010.

\bibitem{632308:15988933}
K.~He, X.~Zhang, S.~Ren, and J.~Sun, ``{Delving Deep into Rectifiers:
  Surpassing Human-Level Performance on ImageNet Classification},''
  \emph{{CoRR}}, vol. abs/1502.01852, 2015.

\bibitem{632308:15988940}
D.~P. Kingma and J.~Ba, ``{Adam: A Method for Stochastic Optimization},''
  \emph{{CoRR}}, vol. abs/1412.6980, 2014.

\bibitem{632308:15988941}
T.-Y. Lin, P.~Goyal, R.~B. Girshick, K.~He, and P.~Doll\'{a}r, ``{Focal Loss
  for Dense Object Detection},'' \emph{{CoRR}}, vol. abs/1708.02002, 2017.

\end{thebibliography}

%




\vskip -2\baselineskip plus -1fil%
\begin{IEEEbiographynophoto}{Rishab Sharma}
    Data Scientist - Fynd Research
\end{IEEEbiographynophoto}

%




\vskip -2\baselineskip plus -1fil%
\begin{IEEEbiographynophoto}{Rahul Deora}
    Data Scientist - Fynd Research
\end{IEEEbiographynophoto}

%




\vskip -2\baselineskip plus -1fil%
\begin{IEEEbiographynophoto}{Anirudha Vishvakarma}
    Principal Engineer - Fynd Research
\end{IEEEbiographynophoto}
\vfill
\end{document}